\documentclass[letterpaper]{article}
\usepackage{aaai}
\usepackage{times}
\usepackage{helvet}
\usepackage{courier}
\usepackage{graphicx}
\newcommand{\headingbf}[1]{\par\vspace{0.5ex}\noindent\textbf{#1}\par\vspace{0.25ex}}
\newcommand{\eqfig}[1]{\par\vspace{0.3ex}\noindent{\centering\includegraphics[width=\linewidth]{#1}\par}\vspace{0.3ex}}

\title{The discovery of the effects of women employment participation on\\the fertility of developing countries: A panel data approach}
\author{Thi Kim Ngan Nguyen\\
Tokyo International University}
\begin{document}
\nocopyright
\maketitle

\begin{abstract}
\begin{quote}
The fertility trend in developing countries experience a significant decline in the last few decades; at the same time, the role of women in the workplace has improved. To have a better insight of the causality of the rate of women participation in the labor market on the total fertility rate in developing world, this paper divides the dataset of 115 developing countries in the period of 1991-2018 into four continents group (Africa, North/South America, Asia/Pacific, Europe) and then applies a data-driven panel data econometric procedure to mitigate omitted bias. The results suggest that the fertility behaviors of women in the North/South America continents are influenced by their career choice; meanwhile in society of other regions, other factors might be more important to women when thinking of having children. In conclusion, policymakers can reference to the paper and formulate policies to have more incentives in making reproductive decisions and further research in the field needs to consider family policies and patrilocality of developing countries as important data.
\end{quote}
\end{abstract}

\section{Introduction}

Recently, the world, especially, many developing countries, has witnessed the fact that fertility has plummeted quickly. More specifically, according to World Fertility Patterns 2015 of the United Nations, the trend of the total fertility rate of less developed regions dropped from 3.4 children per woman in the 1990-1995 period to 2.6 births in 2010-2015, and that is projected to slightly decrease to 2.5 births in the late 2020s. Total fertility rate per woman (TFR) refers to ``the total number of children born or likely to be born to a woman in her lifetime if she were subject to the prevailing rate of age-specific fertility in the population.'' (World Health Organization, 2020). In other words, a woman on average decides to have roughly more than two children, compared to the number of 3 in the previous period (1990-1995). Thus, the context and causes of the change became a concern that raised attention in researchers.

In another context, as female is empowered and encouraged to take part in many social activities, the participation rate of women in the labor force is necessary to be brought into consideration. The U.S. Bureau of Labor Statistics (2020) defines the labor force participation rate as ``a measure of the percentage of a country's working-age population, which usually refers to the age group of 16-64, that engages actively in the labor market, either by currently being employed or by seeking for jobs''. Similarly, women's labor force participation rate refers to the percentage of the female population in the labor market currently working or looking for work.

In the last century, women's participation in the labor force of the age group 16-64 rose significantly across the globe (Tzvetkova, n.d.). Though many countries around the world shared the common pattern of the total fertility rate, the trends of female labor force participation rates are different between high-income and developing countries. In the period from 1990 to 2018, the average rate of women's participation in the developed countries increased by about 4\%, whereas that of the developing world experienced a decrease of 5\% (International Labor Office, 2018). As a result, many researchers have discovered whether there is a causal relationship between TFR and the women's participation rate in the job market in developed countries, meanwhile, there were not an adequate number of studies that cover that of developing countries. This might be needed to be on the top of the agenda since a low fertility rate is a worrisome concern to the developing world.

\section{Background and research motivation}

\subsection{Relationship between fertility and women labor participation in developed and developing countries}

There were numerous of papers discussing about the relationship between the trends of female participating in the labor market and that of fertility, as well as discovering other factors that may influence the fertility. They have different conclusions due to the distinctive approaches and their focused geographical areas. However, the gap between the number of papers mentioning about developing countries and that of developed countries is considerable. As the system of data in OECD countries is well-organized and developed, it is more accessible to researchers who are interested in this country group. On the other hand, when the declining fertility rate in developing world is raising, it is essential to determine whether the changes in the participation rate of women in the workforce contributes to the cause of fertility transition of countries at this level.

Generally, many studies (Emara, 2016; Da Rocha \& Fuster, 2006; Ahn, \& Mira, 2002) have found that there is a significant impact of women's employment rate on fertility as well as discussed some factors that also contribute to the trends of fertility rate in the world. Since the homogeneity characteristics of the development level of nations and data availability play an important role in the empirical study analysis process, researchers frequently concentrated on specific country groups, for instance, developing and developed countries, in order to avoid certain problems (e.g. omitted variable bias). By using data of 29 developing countries from 1990 to 2011, Emara (2016) has done an estimation of the effect of female labor force participation on the fertility rate and illustrated the findings that when the proportion of women taking part in the labor market increases, fertility trends are negatively affected; however, over time, this non-positive impact is diminishing. Meanwhile, when it comes to OECD countries, Da Rocha and Fuster (2006), after observing a positive association between fertility and employment, designed a quantitative theory of decisions of taking part in the labor market and fertility to examine how the labor market frictions play a role in generating this association. By assuming that their jobs being interrupted consequently leads to professional skills reduction, they explored that the low probability of finding jobs possibly makes women's participation rate positively associated with the fertility rate. From a distinct perspective, Ahn and Mira (2002) focus on examining the correlation between TFR and the rate of participation in the labor workforce by applying the Butz-Ward fertility model to their cross-sectional panel data of OCED countries from 1970 to 1995. This model indicated that the fertility response would be procyclical when the employment rate of men increases; but then becomes counter-cyclical as the female employment rate begins to go up due to their opportunity cost of childbearing (wages). Besides that, Ahn and Mira (2002) also discuss the income effects of increasing female wage, the variables of working-hour inflexibility of employees, market child care services, and unemployment ratio. During the research, they show the income and substitution effects of women's income variation by assuming ``children'' as consumer goods. Additionally, when childcare services can be obtained by purchasing, reproductive intentions will also be affected.

Although unemployment has a positive relationship with fertility, when unemployment gets significantly lower in some countries, the fertility case gets even worse.

Though they flexibly applied distinct techniques, Emara (2016) and Ahn \& Mira (2002) in some aspects, agreed that when the women's participation rate in the labor force goes up, the fertility rate declines. On the other side, the results of Da Rocha and Fuster (2006) shows the opposite. As a consequence, the insignificant data size of developing countries, and the conflict between findings in developed countries are points that make us conduct the research.

\subsection{Other determinants on fertility}

When making decisions regarding giving birth, women usually concern not only about their career development but also about other social and environmental factors surrounding them. As a result, in terms of confirming the impact of female employment on fertility, it is inevitable to embrace the importance of other factors in the models. In addition, the Department of Economic and Social Affairs of United Nations (2000) explored that if the continuation of patterns of marriage age, contraception use, level of education, and urbanization will lead to the decrease in fertility by analyzing the reproductive trends and their relatable variables in 74 countries in a 5-year period from 1995-2000.

\headingbf{Family policies}

Many papers (Luci-Greulich and Th\'{e}venon, O., 2013) focus on exploring the effectiveness of government policies regarding family planning by analyzing their effects on fertility. Besides the findings of female employment in developing countries, Emara (2016) also brought up a suggestion that when the duration of paid leave for working mothers and breastfeeding coverage, as well as better compensation, can help improve the trend of fertility. Additionally, in the paper written by Bongaarts (2006), the findings revealed no relationship between a stall of fertility and socioeconomic development trends. Though Bongaarts's discussion considers multiple interesting perspectives, the sample size is small, only acquires seven countries in some experiments.

\headingbf{Educational attainment}

As the gender gap gradually becomes less narrow, the advanced accumulation of education of women also grasps the interest of researchers on the way to seeking the causes of fertility. There is a sociological theory of fertility mentioning that when women have more chance to approach schooling and global networks, they can have a different mindset and desire of family size as they are more aware of planned parenthood and contraception methods (Pradhan, 2016, slide 4). Nevertheless, it is more vital for researchers to understand better how close the relationship between education level and fertility can be. In terms of long-run fertility trends, Lee (2020) indicates that the increase of female educational attainment in the level of primary and secondary reveals a negative impact on fertility; whereas when the fact that more women reach the level of tertiary education, seems to improve the fertility rates. As a consequence, the findings highlight that women's human capital accumulation is vital in the human demographic transition. Therefore, it is reasonable to include the education-related variable in our model.

\headingbf{Political system}

In the realistic world, politics and demography have unavoidable connections because of their adaptability based on the social context of the regions. More specifically, some researchers examine how the fertility rate exerts its impact on the political structure. For instance, Sommer (2018) explores that there is a negative association between democracy and the fertility rate. The findings, which mention that family structure, political and economic status of females are the main reasons for the relationship, were discovered after making comparisons between 140 countries over 30 years. The discussion of this study is significantly considerable because they have collected large-scale data over time and are well-controlled of the data limitations and the reverse causality problem.

\headingbf{Cultural factors}

It is undeniable that fertility transitions usually go along with cultural changes. In the journal article ``Why fertility changes'' written by Charles Hirschman (1994), he discovered that there were a significant number of historical pieces of evidence proving that the distinction between cultural and geographical factors brought more prosper and diverse exploration regarding the origins, pace, and correlation of decreased pattern of childbearing, compared to the projection of a plain theory (pp.203-233). The discussion of this study was surrounding the differences between the reasons of the industrial world's fertility trend in the 1870-1830 period and that of the recent trends in developing countries, which can be dissected into more customized variables. The kinship of household has presented its impact on fertility preferences in a way that patrilocality and fertility norms play main roles in the society, especially developing countries (Grogan, 2012; Bryant, 2002).

\headingbf{Economic growth and human capital investment}

The development of the national economy has a direct relationship with human capital and consequences, it affects the labor supply. In the previous century, the economic growth has been found to have an impact on the fertility (Rouyer, 1987; Perotti, 1996). And likewise, recently, Yujie, L. (2016) has discovered that when human capital investment receives more concentration, the rate of giving birth decreases while the economy develops at a faster pace. Thus, it is necessary to consider the economic growth could play a role in controlling the relationship between TFR and female participation in the labor workforce of the developing world.

In essence, many latter papers discussed various aspects surrounding the TFR and its possible determinants. Nevertheless, TFR of developing countries in the latest decades have not received much attention, and the concern that whether the female employment percentage in these countries have any relations to it has not been discovered. Therefore, we desire to explore more about the effects of the participation of women in the labor market on the TFR while having the cultural and social factors as a control group.

\section{Data}

\subsection{Total fertility rate (TFR)}

In this paper, we create panel data collecting from reliable sources to build a model in discovering the causality of female participation in the labor workforce on the total fertility rate of developing countries. In order to examine whether the participation of women in the job market has a causal effect on the fertility rate (TFR), we choose TFR as the dependent variable, female labor participation rate as the target variable, and other variables related to social, economic and cultural factors as the control variables. Regarding the control group of predictors, it consists of these following variables: female primary education attainment, female tertiary education attainment, GDP growth rate, the percentage of couples living with parents of husbands or wives, and voice and accountability index.

In this study, TFR is the dependent variable. We collected the data from World Bank and original sources are from United Nations. Total fertility rate represents the number of children that would be born to a woman if she were to live to the end of her childbearing years and bear children in accordance with age-specific fertility rates of the specified year. However, there are some limitations and exceptions with the collected data. Annual data series from United Nations Population Division's World Population Prospects are interpolated data from 5-year period data. Consequently, they may not reflect real events as much as observed data.

\subsection{Female labor participation rate (Target variable)}

The data was collected from World Bank and original sources are from United Nations. It refers to the percentage of the population above 15-year-old, who are labor supply for manufacturing goods and services during a period of time. Regarding its limitations, the resulting data often differ from labor force survey data and vary considerably by country, depending on the census scope and coverage. Data provided only on the employed population, not unemployed workers, workers in small establishments, or workers in the informal sector, calculations may systematically over-or underestimate actual rates.

\subsection{Social and environmental variables}

\headingbf{Tertiary education attainment of women} This data is collected from the UNESCO Institute for Statistics. The ratio is calculated without taking ages into consideration and based on the assumption that the target group included people who completed the secondary level of education and meet the requirements to study higher levels. On the other hand, enrollment indicators are based on annual school surveys but do not necessarily reflect actual attendance or dropout rates during the year. Also, the enrollment rates are different from country to country could be due to the distinctiveness of the education systems.

\headingbf{Primary education attainment of women} The data source is from UNESCO Institute for Statistics. It shows the total enrollment to the population acquiring primary education level. Generally, students at this stage are provided with basic knowledge of reading, writing, and mathematics skills as well as other social and natural science subjects.

\headingbf{GDP growth rate} GDP growth rate represents the development of a country compared to the previous year. It is calculated based on the national currency and can be referred to as the growth rate of GDP changes annually. The data source is from World Bank.

\headingbf{The percentage of couples living with parents of husbands or wives} The Global DataLab of Institute for Management Research (Radboud Uni -- Netherlands). The dataset is aggregated from household data of countries having low and middle income. To serve the purpose of the paper, we have combined the percentage of households where couples live with parents of the husband and that of households where couples live with parents of wife in each country by summing them up.

\headingbf{The Voice and Accountability index} The index is provided by the Economist Intelligence Unit. The Voice and Accountability index (VA) is a combination of the democracy index, vested interests, accountability of public officials, human rights, and freedom of association. We use this data to represent the political component of countries.\footnote{Since 2006, VA is a weighted average where the democracy index alone is equally weighted to the four other VA indices. For countries where the other four indices are not given, only the democracy index makes up the VA variable.}

\section{Methodology}

In econometrics, panel data is considered as multi-dimensional data where time-series and cross-sectional data are joined in the panel (Maddala, 2001). Panel analysis has been popular in econometrics studies since it helps explain the changes in trends over time and has the ability to control endogeneity. In this kind of longitudinal data, the fixed effects refer to the representative of the constant changes within observations of the same countries. More specifically, fixed-effects model is a kind of statistical models, which have assumption that the values of predictors are fixed and the dependent variable depends on the values of those predictors.

To grasp better understandings of the effects, we build models with three approaches: the benchmark model, model with variables selected based on prior research, and model with variables selected by double LASSO selection. Firstly, benchmarking is important to examine the efficiency of model. Secondly, prior researches have provided with valuable findings that show many relatable factors of fertility in each continent. Thus, it is essential to have a reference on those papers for variable selection. The final model that is introduced in this work would include control variables that are selected by a method called double LASSO selection. This method was introduced by Urminsky, Hansen, and Chernozhukov (2016) with the purpose of reducing the Type I errors and discover which covariates to imply after two steps. The first step is picking variables that can use for dependent variable prediction and the second step is choosing variables to predict independent variables. The advantage of this method is that it minimizes the overfitting and bias problem and is useful to identify the predictor acquiring enough empirical data to analyze the relationships between variables.

\[
y_{it} = \alpha + \beta x_{it} + \mu_i + \varepsilon_{it}
\]

$y$ is the dependent variable

$x$ is the independent variable that is chosen to examine the causal effect on $y$

$\alpha$ is the intercept (which is constant)

$\beta$ is the coefficient of the independent variable

$\mu$ is the variable that is unobservable

$\varepsilon$ is the residual (error term)

$i$ is the entity's notation (e.g. geographical factors)

$t$ is the notation of time (e.g. year)

*Because of the assumption that $\mu$ is constant over time, the ``i'' subscript is unnecessary.

\section{Empirical Analysis}

The study is conducted by employing the panel data approach, double LASSO selection method to select variables. More specifically, we use the panel data of 115 developing countries in the period of 1991-2018 to understand more about the causal relationship between fertility and its potential determinants. We divided countries into four region-based groups as follows: Africa, South/North America, Asia/Pacific, and Europe groups, to examine the impact of underlying social and cultural factors. To choose the most valuable explanatory variable for the model, we apply a modern technique in Machine learning which is a double LASSO regression model. After the variable selection part, we then use those chosen variables for each group divided as above in the fixed or pooled models consecutively.

In order to examine the causal effect of labor participation of females on the TFR, this study implies the model as follow as a benchmark:
\eqfig{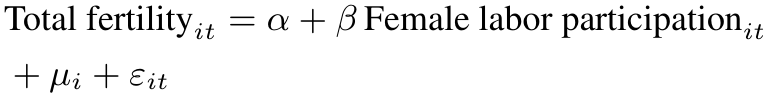}

The fixed-effects model (FEM) was used for 4 regions (Africa, Americas (North/South), Asia/Pacific, and Europe) (see Appendix A for countries in each regions). This model includes the dummy variable of countries ($\mu_i$) representing unobserved distinctive factors for each country (i.e. family policies, minimum wages, and healthcare facilities). Generally, we run the model on 4 groups of regions respectively in order to control their differences.

Continuously, control variables regarding social, economic, and cultural factors are useful to discover whether they have any impacts on the TFR and the labor participation of women. More specifically, we imply the rate of tertiary education attainment of women, GDP growth rate, the proportion of couples living with their parents, and the political system of countries. The foundation of choosing those variables is the discoveries of previous research papers (Pradhan, 2016; Lee, 2020; Sommer, 2018; Hirschman, 1994). To present a better grasp from the political perspective, we use the VA index, instead of taking advantage of democracy only.

\subsection{Analysis}

\headingbf{Africa}

There is a discovery that the higher education level of women leads to the lower fertility in the sub-Saharan Africa and other countries in the Africa region (Shapiro, 2012; Kravdal, 2012). Shapiro (2012) also found that the differences of fertility increases when the level of adjacent education groups increases relative to uneducated women. When it comes to political system, throughout recent decades, the continent has witnessed ups and downs and noble changes of democracy transitions. As a result, the changes in reproductive health system and laws of women rights were improved and found to acquire certain success, notably South Africa (Cooper et al, 2016). However, some remained socio-economic factors such as gender inequalities in income may bring conflict to the positive changes of politics, were also taken into account. Recently, the negative correlation between TFR and GDP per capita of Sub-Saharan Africa has been discovered (Gotmark \& Andersson, 2020), which suggested us to have the GDP growth into consideration. In conclusion, we choose the control variables as following: VA democracy, percentages of female with primary education, percentages of female with tertiary education, and GDP growth. The Africa group has the regression as below:
\eqfig{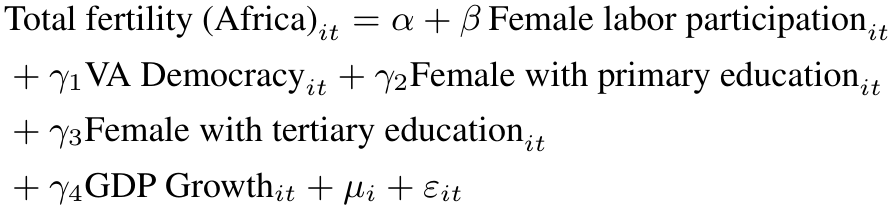}

After following steps of the method called double selection LASSO, we have the regression with chosen variables of the Africa group as follow:
\eqfig{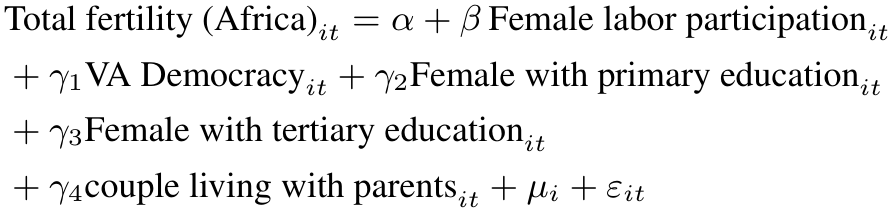}

\begin{figure}[t]
\centering
\includegraphics[width=\linewidth]{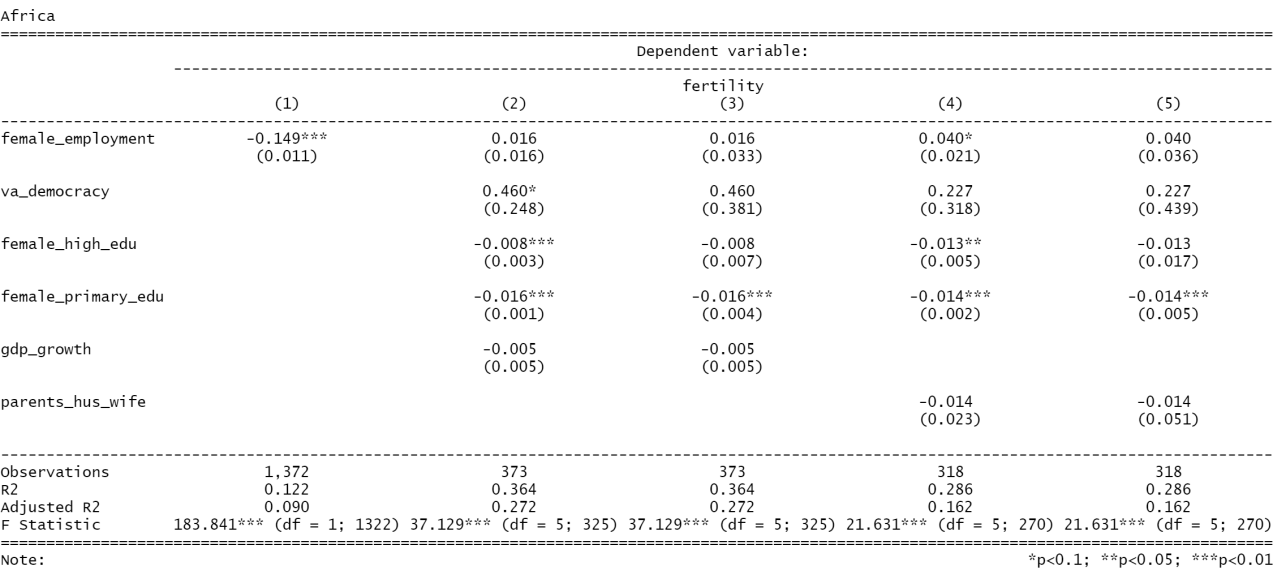}
\caption{Africa. (1) Bench mark; (2) Manual selected variables without robust SE; (3) Manual selected variables with robust SE; (4) Double LASSO selected variables without robust SE; (5) Doubles LASSO selected variables with robust SE}
\label{tab:africa}
\end{figure}

In the table 1, it is shown that the variable of female participation in the labor market implies a significant negative effect on the TFR of Africa. However, since the bench mark model does not cover other factors, this should not be reliable to interpret. Accordingly, the model which we chose variables manually and the one we use the double LASSO selection method indicates that the impact of female participation in the workforce of Africa is statistically insignificant and thus can be implied as there is no effect of the target variable on the outcome variable. In essence, there is no effect of the female participation in the labor workforce on fertility rate in the Africa continent.

\headingbf{South/North America}

South/North America have been through plenty changes in the period of 2 decades, 1991-2018. Because of that, people desire to expand their knowledge of reasons regarding the fertility transitions during that time. Difference in social status and education led to the variety in fertility choices among populations in North Americas (Dribe et al., 2017). Additionally, the economic context in Latin America has been found to have a positive effect on fertility and women who has the access to education tend to postpone their maternity (Adsera and Menendez, 2013). Thus, we decided to include the female tertiary education and primary education attainment, and GDP growth rate into the manual model. . In conclusion, the chosen control variables for this group are: percentages of female with primary education, percentages of female with tertiary education, and GDP growth. The South/North America group has the regression as below:
\eqfig{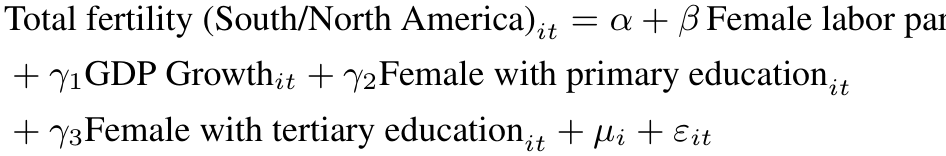}

After following steps of the method called double selection LASSO, we have the regression with chosen variables of the South/North America group as follow:
\eqfig{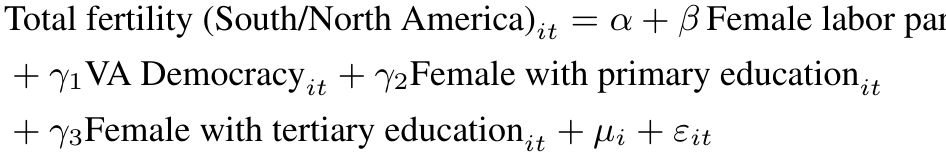}

\begin{figure}[t]
\centering
\includegraphics[width=\linewidth]{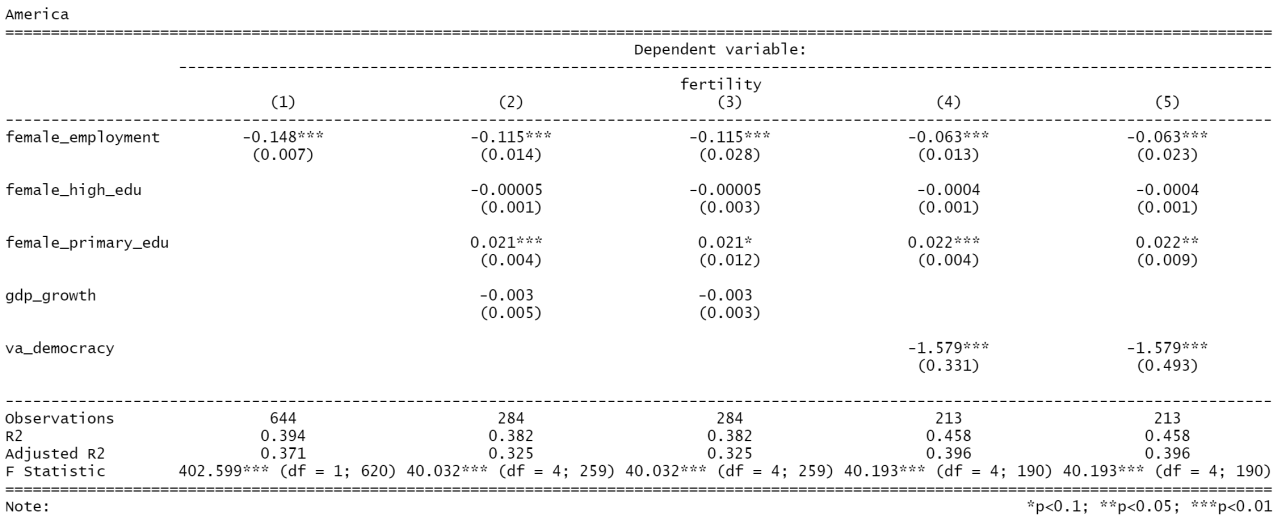}
\caption{South/North America. (1) Bench mark; (2) Manual selected variables without robust SE; (3) Manual selected variables with robust SE; (4) Double LASSO selected variables without robust SE; (5) Doubles LASSO selected variables with robust SE}
\label{tab:america}
\end{figure}

In the case of America, we observe that the participation of women in South/North America is significant in all proposed models. However, the negative relationship of the female participation in the labor market with the total fertility is overestimated in the benchmark model.

\headingbf{Asia/Pacific}

In the Asia/Pacific area, the fertility trends depend considerably on the cultural and educational factors. The education improvement among women and childbearing cost have been found to reduce fertility in some Asian countries (Caldwell, 1982). Besides that, household context plays a role in fertility decline recently since traditional cultural perspective, such as son preference and family kinship, have transformed throughout the years (Chan and Yeoh, 2002). In other words, the family attachment with previous generations could lead to the number of children the couple decide to have. Thus, along with the education-related predictors, the value representing the number of couples living with parents is also decent to put into the model. In conclusion, the picked control variables are: percentages of female with primary education, percentages of female with tertiary education, and couple living with parents. The Asia/Pacific group has the regression as below:
\eqfig{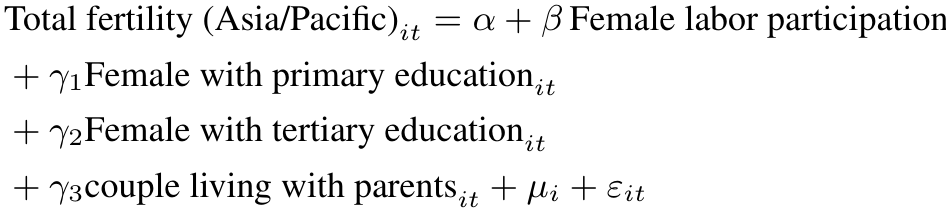}

After following steps of the method called double selection LASSO, we have the regression with chosen variables of the Asia/Pacific group as follow:
\eqfig{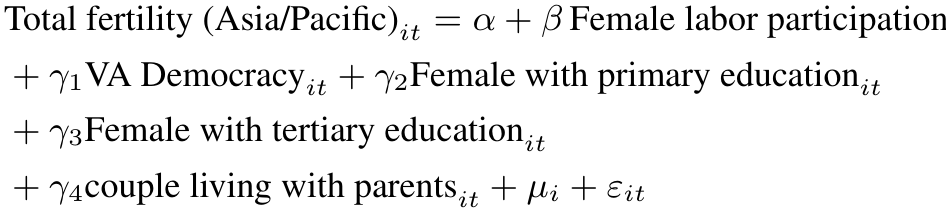}

\begin{figure}[t]
\centering
\includegraphics[width=\linewidth]{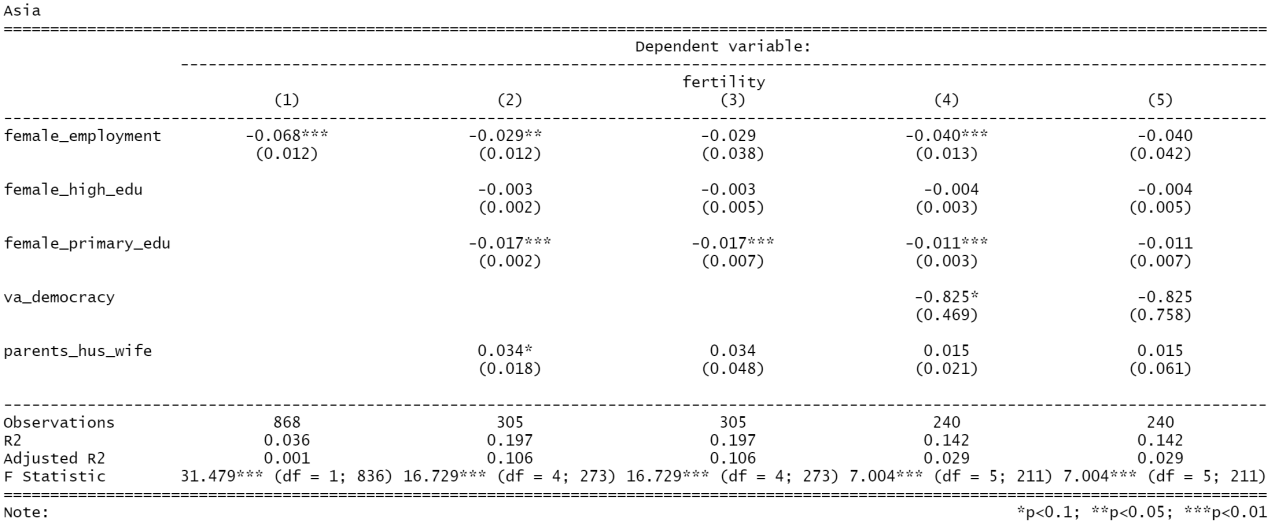}
\caption{Asia/Pacific. (1) Bench mark; (2) Manual selected variables without robust SE; (3) Manual selected variables with robust SE; (4) Double LASSO selected variables without robust SE; (5) Doubles LASSO selected variables with robust SE}
\label{tab:asia}
\end{figure}

The observations from the Asia/Pacific region suggest that our target variable is significant in the benchmark model and the double LASSO selection implied FEM. There is a negative effect of the female participation in the job market on the TFR of Asia when we do not control homoskedasticity. Meanwhile, with the model (5) having the robust SE, the target variable becomes insignificant, which could due to the variation of data. Generally, it can be said that the employment participation of female does not show significant effect on the TFR.

\headingbf{Europe}

The rapid decline in TFR of European countries has raised concerns in researchers for the last decades. The employment of women and the fluctuations of business context in Europe were revealed to be associated with the behaviors of fertility. On the side of family policies, there was a exploration in 1980s that concluded the political system and the reluctance of policy markers contributes to the dramatic fall in Western Europe (McIntosh, 1981). In the material of UNESCO called ``Adult and youth literacy: National, regional and global trends, 1985--2015, shown data reveals that the literacy rate of adults in Europe countries did not experience any noticeable changes, which means the primary education system seems to be consistent. Therefore, the variable of female acquiring primary education was not significant in contributing to the trend of fertility. Instead, the increase of highly educated women tends to be more vital in the postponement of reproductive decisions in this continent. Testa (2014) presents that the association between the education attainment and choices to have children is positive at both micro and macro levels after analyzing survey data of 27 European countries. In conclusion, we choose the control variables as following: VA democracy, percentages of female with tertiary education, and GDP growth. The Europe group has the regression as below:
\eqfig{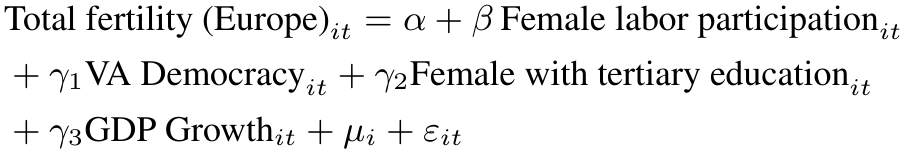}

After following steps of the method called double selection LASSO, we have the regression with chosen variables of the Asia/Pacific group as follow:
\eqfig{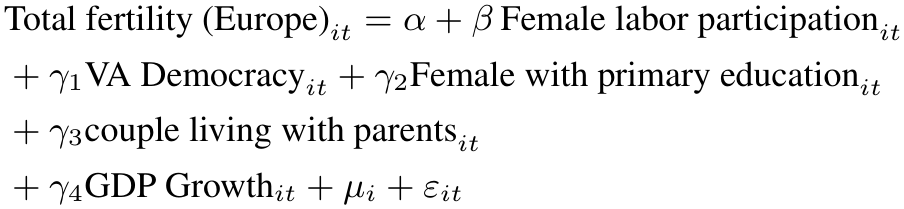}

\begin{figure}[t]
\centering
\includegraphics[width=\linewidth]{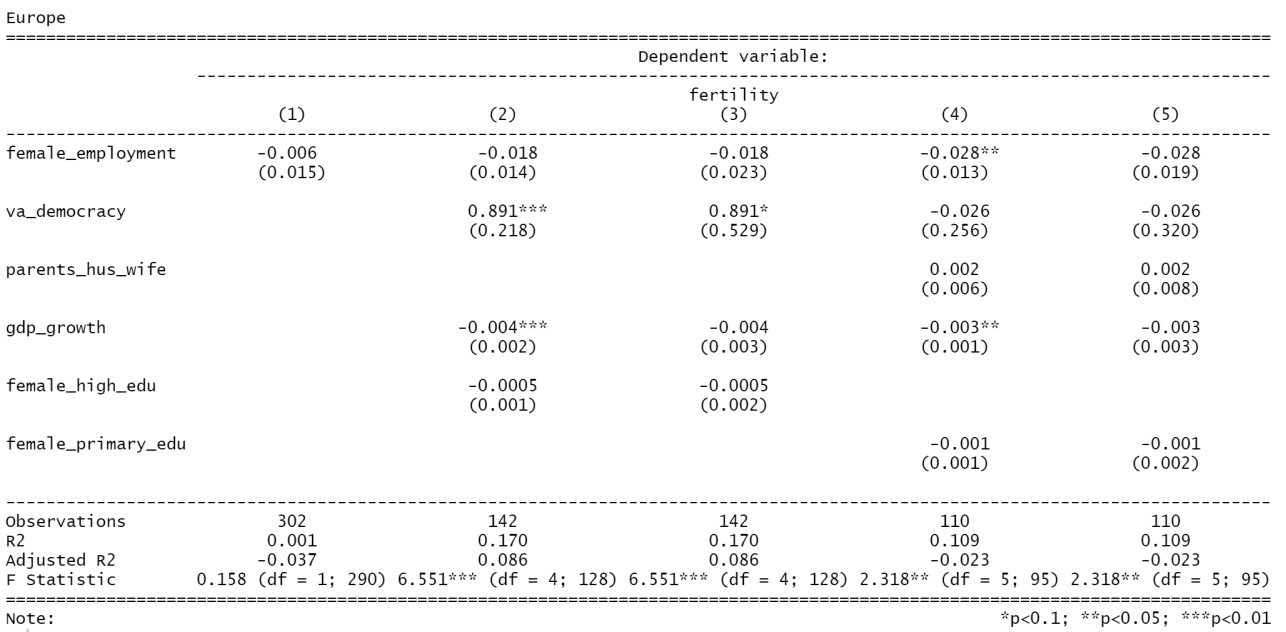}
\caption{Europe. (1) Bench mark; (2) Manual selected variables without robust SE; (3) Manual selected variables with robust SE; (4) Double LASSO selected variables without robust SE; (5) Doubles LASSO selected variables with robust SE (*). *All the models have the fixed effects models}
\label{tab:europe}
\end{figure}

Regarding Europe's developing countries, the female employment participation is shown to have no significant effect on the total fertility. In the model (4), there is a significance under 5\% of the target variable, but when the homoskedasticity is under controlled, which is the result of the model (5), it became insignificant. This shows that in Europe, the fertility is not influenced by the female employment participation.

\section{Conclusion and discussion}

The decline of fertility rate around the world has become a worrisome concern, which has been a topic that researchers desire to learn more about its causes. It is undeniable that there are numerous papers mentioning the relationship between the childbearing and female participation in the labor market within the scope of developed nations, more specifically OECD countries. On the other hand, in the developing world, the materials providing points of view about the decreasing trend of fertility are insufficient, due to the limitation of data and the complexity of cultural aspects regarding its possible determinants. Therefore, this paper intents to analyze whether the impact of women involving in the workplace on their decision making for childbirth is significant in emergent regions.

In order to explore an approximation of the causal effect of female participation in the labor workforce on the TFR in the developing countries, we use empirical approach by applying 3 models with panel data. The dataset consisting national data of 115 countries in the period of 1991-2018 and was divided into 4 regional groups: Africa, America, Asia/Pacific and Europe. The result illustrates that the number of women taking part in the labor market have negative impact on the fertility rate of the South/North America continents. Meanwhile, we found no significant evidence of the relationship between the patterns of women labor participation and the birthrate in three other continents, Africa, Asia/Pacific and Europe. This can be inferred that the fertility choices of women in the South/North America continents are influenced by their career choice; meanwhile in society of other regions, other factors might be more important to women when thinking of having children. Though we have limited the omitted variable bias, some important factors related to policies have not been brought into the models due to the limitation of data. Thus, further research needs to consider family policies and patrilocality of developing countries. The topic of childbearing and working trade-off that women have to face in developing countries need to be paid more attention. Additionally, the paper can be considered a reference for policymakers to formulate family policies to have more incentives in making reproductive decisions.

\section*{References}

Acemoglu, D., \& Autor, D. (2011). \textit{Lectures in labor economics}. Manuscript.

Ahn, N., \& Mira, P. (2002). A note on the changing relationship between fertility and female employment rates in developed countries. \textit{Journal of Population Economics}, 15(4), 667-682.

Bongaarts, J. (2008). Fertility transitions in developing countries: Progress or stagnation?. \textit{Studies in family planning}, 39(2), 105-110.

Butz, W. P., \& Ward, M. P. (1979). The emergence of countercyclical US fertility. \textit{The American economic review}, 69(3), 318-328.

Bryant, J. (2002). Patrilines, patrilocality and fertility decline in Viet Nam. \textit{Asia-Pacific Population Journal}, 17(2), 111-128.

Caldwell, J. C. (1982). Theory of fertility decline. academic press.

Chan, A., \& Yeoh, B. S. (2002). Gender, family and fertility in Asia: An introduction. \textit{Asia-Pacific Population Journal}, 17(2), 5-10.

Cooper, D., Harries, J., Moodley, J., Constant, D., Hodes, R., Mathews, C., ... \& Hoffman, M. (2016). Coming of age? Women's sexual and reproductive health after twentyone years of democracy in South Africa. \textit{Reproductive health matters}, 24(48), 79-89.

Da Rocha, J. M., \& Fuster, L. (2006). Why are fertility rates and female employment ratios positively correlated across OECD countries?. \textit{International Economic Review}, 47(4), 1187-1222.

Dribe, M., Breschi, M., Gagnon, A., Gauvreau, D., Hanson, H. A., Maloney, T. N., ... \& V\'{e}zina, H. (2017). Socio-economic status and fertility decline: Insights from historical transitions in Europe and North America. \textit{Population studies}, 71(1), 3-21.

G\"{o}tmark, F., \& Andersson, M. (2020). Human fertility in relation to education, economy, religion, contraception, and family planning programs. \textit{BMC Public Health}, 20(1), 1-17.

Grogan, L. (2013). Household formation rules, fertility and female labour supply: Evidence from post-communist countries. \textit{Journal of Comparative Economics}, 41(4), 1167-1183.

Hausman, J. A. (1978). Specification tests in econometrics. \textit{Econometrica: Journal of the econometric society}, 1251-1271.

Hirschman, C. (1994). Why fertility changes. \textit{Annual review of sociology}, 203-233.

International Labour Office. (2018, July). ILO labour force estimates and projections (LFEP) 2018.

Lee, J. W. (2020). Determinants of fertility in the long run. \textit{The Singapore Economic Review}, 65(04), 781-804.

Li, Y. (2016). The relationship between fertility rate and economic growth in developing countries.

Luci-Greulich, A., \& Th\'{e}venon, O. (2013). The impact of family policies on fertility trends in developed countries. \textit{European Journal of Population/Revue europ\'{e}enne de D\'{e}mographie}, 29(4), 387-416.

Maddala, G. S. (2001). Introduction to econometrics, John Wiley and Sons. West Sussex, England.

McIntosh, C. A. (1981). Low fertility and liberal democracy in Western Europe. \textit{Population and Development Review}, 181-207.

Perotti, R. (1996). Growth, income distribution, and democracy: What the data say. \textit{Journal of Economic growth}, 1(2), 149-187.

Pradhan, E. (2016). \textit{Link between education and fertility in low and middle income countries} [PowerPoint slides]. United Nations.

Przeworski, A., Alvarez, R. M., Alvarez, M. E., Cheibub, J. A., Limongi, F., \& Neto, F. P. L. (2000). Democracy and development: Political institutions and well-being in the world, 1950-1990 (No. 3). Cambridge University Press.

Rouyer, A. R. (1987). Political capacity and the decline of fertility in India. \textit{American Political Science Review}, 81(2), 453-470.

Shapiro, D. (2012). Women's education and fertility transition in sub-Saharan Africa. \textit{Vienna Yearbook of Population Research}, 9-30.

Sommer, U. (2018). Women, demography, and politics: how lower fertility rates lead to democracy. \textit{Demography}, 55(2), 559-586.

Testa, M. R. (2014). On the positive correlation between education and fertility intentions in Europe: Individual-and country-level evidence. \textit{Advances in life course research}, 21, 28-42.

Tzvetkova, E. (n.d.). Working women: Key facts and trends in female labor force participation. Retrieved October 16, 2020.

UNESCO Institute for Statistics. (2013). Adult and youth literacy: National, regional and global trends, 1985--2015. Montreal: UNESCO Institute for Statistics.

United Nations. (2000). \textit{Fertility levels and trends in countries with intermediate levels of fertility}. Retrieved October 10, 2020.

United Nations. (2015). \textit{World fertility patterns 2015}. Retrieved October 10, 2020.

Urminsky, O., Hansen, C., \& Chernozhukov, V. (2016). Using double-lasso regression for principled variable selection. Available at SSRN 2733374.

U.S. Bureau of Labor Statistics. (n.d.). \textit{Labor force statistics from the current population survey}. Retrieved September 12, 2020.

World Health Organization. (n.d.). \textit{Health situation and trend assessment}. Retrieved September 12, 2020.

\section*{Appendix A}

List of countries in the dataset:

1) Africa: Angola, Burundi, Benin, Burkina Faso, Botswana, Central African Republic CAR, Cote d'Ivoire, Cameroon, Congo Democratic Republic, Congo Brazzaville, Comoros, Djibouti, Algeria, Egypt, Eritrea, Ethiopia, Gabon, Ghana, Guinea, Gambia, Guinea, Gambia, Guinea Bissau, Kenya, Liberia, Lesotho, Morocco, Madagascar, Mali, Mozambique, Mauritania, Malawi, Namibia, Niger, Nigeria, Rwanda, Sudan, Senegal, Sierra Leone, Somalia, South Sudan, Sao Tome \& Principe, Eswatini, Chad, Togo, Tunisia, Tanzania, Uganda, South Africa, Zambia, Zimbabwe (Total: 49)

2) North/South America: Argentina, Belize, Bolivia, Brazil, Barbados, Chili, Colombia, Costa Rica, Cuba, Guatemala, Guyana, Honduras, Haiti, Jamaica, Saint Lucia, Mexico, Panama, Peru, Paraguay, El Salvador, Suriname, Trinidad \& Tobago, Uruguay (Total: 23)

3) Asia/Pacific: Afghanistan, Bangladesh, Bhutan, China, Indonesia, India, Iran, Iraq, Jordan, Kazakhstan, Kyrgyzstan, Cambodia, Lao, Maldives Myanmar, Mongolia, Nepal, Pakistan, Philippines, Papua New Guinea, Palestine, Qatar, Thailand, Tajikistan, Turkmenistan, Timor-Leste, Turkey, Uzbekistan, Vietnam, Vanuatu, Yemen (Total: 31)

4) Europe: Albania, Armenia, Azerbaijan, Bosnia Herzegovina, Belarus, Georgia, Moldova, North Macedonia, Monte Negro, Serbia, Ukraine, Kosovo (Total: 12)

\end{document}